\documentclass[sigconf]{acmart}
\usepackage{colortbl}
\usepackage{microtype}
\usepackage{balance}
\usepackage{tabularx}
\usepackage{float}
\AtBeginDocument{%
  }

\copyrightyear{2026}
\acmYear{2026}
\setcopyright{cc}
\setcctype{by}
\acmConference[KDD 2026] {Proceedings of the 32nd ACM SIGKDD Conference on Knowledge Discovery and Data Mining V.2}{August 9--13, 2026}{Jeju Island, Republic of Korea.}
\acmBooktitle{Proceedings of the 32nd ACM SIGKDD Conference on Knowledge Discovery and Data Mining V.2 (KDD 2026), August 9--13, 2026, Jeju Island, Republic of Korea}
\acmISBN{979-8-4007-2259-2/2026/08}
\acmDOI{10.1145/3770855.3817564}
\begin{document}

\title{SVHighlights: Towards Extremely Long Sport Video Highlight Detection}

\author{Donggyu Lee}
\authornote{Both authors contributed equally to this research.}
\orcid{0009-0008-6887-582X}
\email{leedongkyu2019@unist.ac.kr}
\author{Youngbin Ki}
\authornotemark[1]
\orcid{0009-0001-9135-209X}
\email{youngbinki@unist.ac.kr}
\affiliation{%
  \institution{Ulsan National Institute of Science and Technology}
  \city{Ulsan}
  \country{Republic of Korea}
}

\author{Jeonghun Kang}
\orcid{0009-0002-8343-4404}
\email{jhkang@unist.ac.kr}
\affiliation{%
  \institution{Ulsan National Institute of Science and Technology}
  \city{Ulsan}
  \country{Republic of Korea}
}

\author{Taehwan Kim}
\orcid{0000-0002-6571-4632}
\email{taehwankim@unist.ac.kr}
\affiliation{%
  \institution{Ulsan National Institute of Science and Technology}
  \city{Ulsan}
  \country{Republic of Korea}
}

\renewcommand{\shortauthors}{Donggyu Lee, Youngbin Ki, Jeonghun Kang, and Taehwan Kim}


\begin{abstract}
While highlight detection for long-form videos is of great practical importance, most existing methods remain limited to short-form content, largely due to the absence of a suitable benchmark. To bridge this gap, we introduce SVHighlights, to the best of our knowledge, the first benchmark for highlight detection in extremely long sports videos, each exceeding one hour in duration, across multiple sports categories. SVHighlights is constructed from pairs of full-length sports videos and their corresponding official highlight videos using a dataset generation pipeline, enabling scalable and cost-effective label generation without conventional per-clip saliency annotation. The benchmark comprises 320 videos spanning a wide range of sports, with an average duration of 2.00 hours and a total of 640.18 hours, substantially exceeding previous highlight detection datasets. Beyond the lack of benchmarks, existing methods also face fundamental challenges on long videos: models trained on short clips of only a few minutes fail to generalize to hour-long content, and their clip-level scoring lacks the broader context needed to identify highlights in long-form videos. To address these challenges and provide a strong baseline for SVHighlights, we present TF-SELECTOR, a training-free segment-based approach that divides each video into context-aware segments by merging adjacent shots sharing the same semantic content, and predicts segment-level saliency scores using a large language model (LLM) with multimodal inputs including visual captions, transcripts, and audio volume. Extensive experiments demonstrate that TF-SELECTOR achieves superior performance across most evaluation metrics compared to Video Temporal Grounding (VTG)-tuned baselines, with improvements of +2.50 in HIT@1, +4.04 in HIT@K, and +2.95 in IoU. These results establish SVHighlights as a challenging testbed for long-form highlight detection and demonstrate that a simple segment-based strategy can effectively scale to hour-long videos. The dataset and code are available at \url{https://leedongkyu2019.github.io/SVHighlights/}.

\end{abstract}    




\maketitle

\begin{figure}[!t]
  \centering
  \includegraphics[width=\columnwidth]{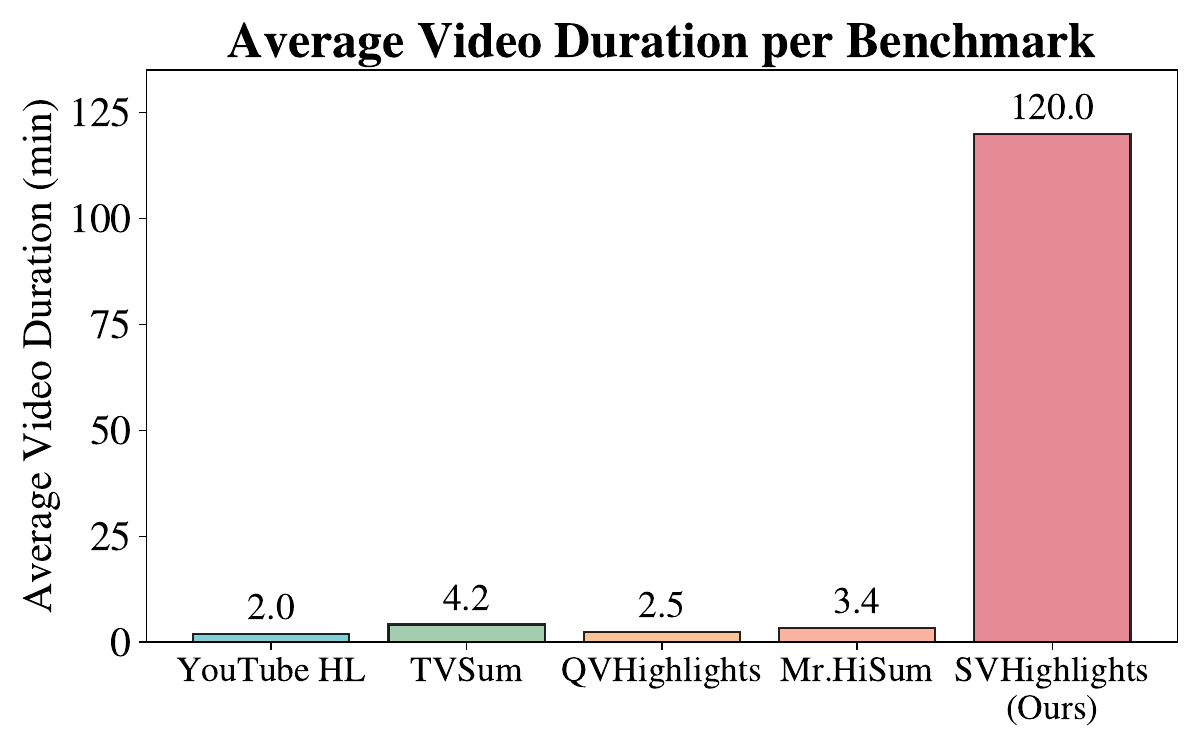}
  \caption{Average video duration (in minutes) for each video highlight detection benchmark dataset. Our SVHighlights dataset contains significantly longer videos on average compared to existing benchmarks, providing a unique testbed for long-form video understanding.
}
  \Description{Bar chart comparing average video duration across five highlight detection benchmarks. SVHighlights shows an average of 120 minutes, far exceeding YouTube Highlights, TVSum, QVHighlights, and Mr.HiSum, which are all under 5 minutes.}
  \label{fig:length}
\end{figure}
\section{Introduction}
\label{sec:intro}

Across major video platforms, viewers increasingly prefer concise, engaging content—such as sports highlights, YouTube Shorts, or recaps of movies and TV shows—over watching full-length videos~\cite{violot2024shorts, guan2024impact}. However, manually extracting highlight-worthy moments from long videos is both time-consuming and costly, rendering large-scale highlight production impractical. Consequently, there has been a growing demand for automatic highlight detection systems~\cite{wang2025long, sul2023mr, kwak2025effective}.

Although many promising methods have been proposed~\cite{lin2023univtg, liu2024r, lei2021detecting, liu2022umt, islam2025unsupervised}, existing research has predominantly focused on short-form videos. A primary reason for this limitation is the absence of a suitable benchmark for highlight detection in long-form content. Constructing such a benchmark is particularly challenging, as most existing datasets depend on manual annotations~\cite{lei2021detecting, song2015tvsum, sun2014ranking}, which are difficult to scale to videos spanning several hours; to annotate highlights, annotators must watch the entire video, making the process prohibitively time-consuming and labor-intensive.

To address these limitations and foster further research, we introduce SVHighlights, to the best of our knowledge, the first benchmark for highlight detection in extremely long sports videos, each exceeding one hour in duration, across multiple sports categories.
We focus on the sports domain for two main reasons. First, sports events contain clearly defined exciting moments (e.g., goals, scores), which provide unambiguous ground truth for highlights. Second, official sports channels on platforms such as YouTube regularly upload full-length match recordings alongside professionally edited highlight videos, making large-scale data collection both practical and reliable.
Our benchmark is constructed from pairs of full-length sports videos and their corresponding highlight videos collected from official YouTube channels. Instead of relying on costly manual annotations, we use the official highlight videos as ground truth. Specifically, we employ a highlight alignment algorithm to automatically identify which segments of a full-length video appear in its corresponding highlight video, enabling scalable label generation across a large corpus of long videos.
In total, we collected 320 videos from YouTube spanning a diverse range of sports categories. The benchmark contains videos with an average duration of 2.00 hours, summing up to a total of 640.18 hours. As illustrated in Figure~\ref{fig:length}, SVHighlights features average video durations that are substantially longer than those in previous highlight detection benchmarks.

Beyond the benchmark gap, existing highlight detection methods face fundamental challenges when applied to long-form content. Most Video Temporal Grounding (VTG) models are trained on short-video benchmarks such as QVHighlights~\cite{lei2021detecting}, where videos average only 2.5 minutes. When applied to hour-long videos, these models struggle to generalize due to the drastically different temporal dynamics and highlight distributions. Moreover, their clip-level scoring approach lacks the broader context needed to determine whether a given moment constitutes a highlight within a lengthy video. Recently, there has been increasing interest in leveraging the reasoning capabilities of large language models (LLMs) for video highlight detection in a zero-shot setting~\cite{ren2024timechat, guo2024trace, guo2025vtg}. However, LLM-based approaches necessitate strict constraints on the number of input frames, leading to substantial information loss for long videos. For instance, if an LLM-based model is limited to processing 96 frames, uniformly sampling a two-hour video would result in a frame every 75 seconds—missing key highlight events entirely.

To address these challenges and provide a strong baseline for SVHighlights, we present TF-SELECTOR (Training-Free Segment-based Extremely Long video highlight detECTOR), a training-free framework that combines off-the-shelf foundation models with a simple segment-based processing strategy. TF-SELECTOR forms context-aware segments by detecting shot boundaries and merging adjacent shots that share semantic content via transcript cues, then predicts a saliency score per segment and assigns it to all of its clips. This segment-level design yields consistent scores across clips depicting the same scene and lets the LLM process each segment with enough frames, addressing the context and frame-sampling limitations of clip-level LLM-based approaches.

We evaluate the effectiveness of TF-SELECTOR on SVHighlights. Experimental results show that our method achieves superior performance across most metrics compared to VTG-tuned baselines, demonstrating that our training-free approach can not only process long videos effectively but also predict saliency scores accurately.

In summary, our main contributions are as follows:
\begin{itemize}
    \item We introduce SVHighlights, the first benchmark for highlight detection in extremely long sports videos, comprising 320 videos with an average duration of 2.00 hours—approximately 30 to 60 times longer than existing datasets. This benchmark addresses a critical gap in evaluating highlight detection methods at scale.
    \item We develop a dataset generation pipeline that aligns official highlight videos with full-length broadcasts, enabling scalable dataset construction that requires only lightweight manual verification instead of labor-intensive per-clip saliency annotation.
    \item We establish TF-SELECTOR as a strong baseline for SVHighlights. This training-free approach demonstrates that combining off-the-shelf vision-language models with segment-based processing can outperform existing VTG-tuned methods on hour-long videos, while also revealing significant room for future improvement.
\end{itemize}
\section{Related Work}
\label{sec:related_work}

\subsection{Video Highlight Detection} Many existing video highlight detection methods~\cite{garcia2018phd, jiao2018video, yu2018deep, rochan2020adaptive, badamdorj2021joint, sun2014ranking, gygli2016video2gif, xu2021cross} are trained on datasets with frame-level annotations, where each frame is manually labeled by human annotators to indicate whether it constitutes a highlight. A few methods such as Jiao et al.~\cite{jiao2018video} and SL-Module~\cite{xu2021cross} instead predict highlight scores at the segment level. Recently, there has been growing interest in combining moment retrieval and highlight detection, where the goal is to extract highlights that are semantically aligned with a given textual query~\cite{lei2021detecting, liu2022umt, moon2023query, xu2024mh, lin2023univtg, sun2024tr, liu2024r}. In addition to these supervised methods, recent studies such as TimeChat~\cite{ren2024timechat}, VTG-LLM~\cite{guo2025vtg}, and TRACE~\cite{guo2024trace} explore using LLMs for highlight detection and demonstrate strong zero-shot performance on the QVHighlights benchmark. However, all of these methods are designed for and evaluated on short-form videos, and it remains unclear how well they generalize to long-form content where temporal dynamics and highlight distributions differ substantially. In contrast, TF-SELECTOR forms context-aware, variable-length segments by merging semantically related shots, extending segment-level processing to long-form content.

\subsection{Long Video Highlight Detection} Manually annotating highlights in long videos is costly and time-consuming, which is why most existing research on long video highlight detection has been limited to the sports domain—where official highlight videos produced by broadcasters are readily available. \citet{shukla2018automatic} proposed a model for cricket highlight generation by combining event-driven and excitement-based approaches. H5~\cite{merler2018automatic} detects highlights in golf and tennis by estimating excitement based on player actions, facial expressions, crowd reactions, and commentator speech. \citet{della2025automated} segment full-length soccer videos into 5-second clips and classify each clip as a highlight or not based on audio and video features. While these works demonstrate the feasibility of long video highlight detection, each targets a single sport and relies on ad-hoc evaluation protocols such as user studies or manually constructed test sets, making it difficult for other researchers to reproduce or compare results across methods. This lack of a standardized benchmark for long-form highlight detection motivates the construction of SVHighlights.

\subsection{Video Highlight Detection Benchmarks} Many existing benchmarks rely on human annotators to assign importance scores, which makes it difficult to scale to long videos. Consequently, prior benchmark studies have primarily focused on short-form videos. The YouTube Highlights dataset~\cite{sun2014ranking} is a widely used benchmark for highlight detection, created by collecting raw and edited videos from YouTube across six domains. For evaluation, the test split is annotated by five human annotators. TVSum~\cite{song2015tvsum} contains 50 YouTube videos across 10 categories, with five videos per category, and each video is approximately 4 minutes long. Each video is annotated by 20 human annotators who assign importance scores to 2-second shots. A representative benchmark for video moment retrieval combined with highlight detection is QVHighlights~\cite{lei2021detecting}, which comprises over 10,000 YouTube videos with an average duration of 2.5 minutes. In this dataset, human annotators were instructed to select 2-second clips relevant to a query, and 3 annotators subsequently assigned a saliency score to each clip. More recently, Mr.HiSum~\cite{sul2023mr} introduced a large-scale highlight detection dataset of 31,892 videos with automatically generated labels, yet its average video duration remains only 3.4 minutes. As summarized in Table~\ref{tab:highlight-benchmarks}, all existing datasets focus on short-form videos, leaving no standardized benchmark for evaluating highlight detection in long-form content.

\begin{table}[t]
\centering
\caption{List of existing video highlight detection datasets and their statistics.}
\label{tab:highlight-benchmarks}
\resizebox{\linewidth}{!}{%
\begin{tabular}{lcccc}
\toprule
Benchmark & \#Vids & Dur. (h) & Avg. (min) & Labels \\
\midrule
YouTube HL~\cite{sun2014ranking} & 712 & 23.8 & 2.0 & Auto/Human \\
TVSum~\cite{song2015tvsum} & 50 & 3.5 & 4.2 & Human \\
QVHighlights~\cite{lei2021detecting} & 10,148 & 422.8 & 2.5 & Human \\
Mr.HiSum~\cite{sul2023mr} & 31,892 & 1,788 & 3.4 & Auto \\
\midrule
\textbf{SVHighlights (Ours)} & \textbf{320} & \textbf{640.18} & \textbf{120.0} & \textbf{Auto(Human)}\protect\footnotemark \\
\bottomrule
\end{tabular}
}
\end{table}
\footnotetext{\textit{Auto(Human): automatic labels with lightweight human trimming and filtering steps (Section~\ref{sec:benchmark}), not per-clip saliency annotation.}}

\begin{figure}[t]
  \centering
  \includegraphics[width=\columnwidth]{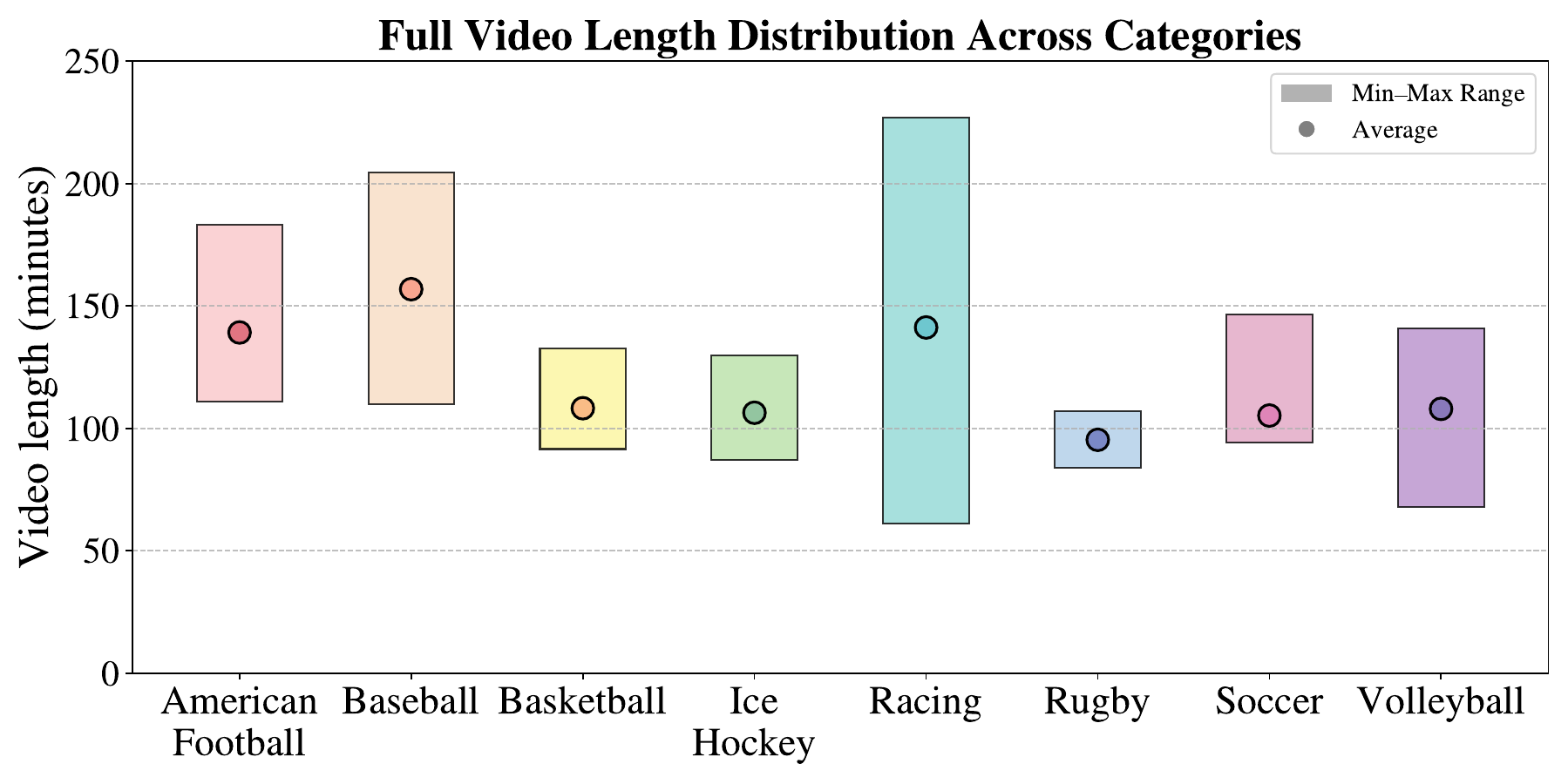}
  \includegraphics[width=\columnwidth]{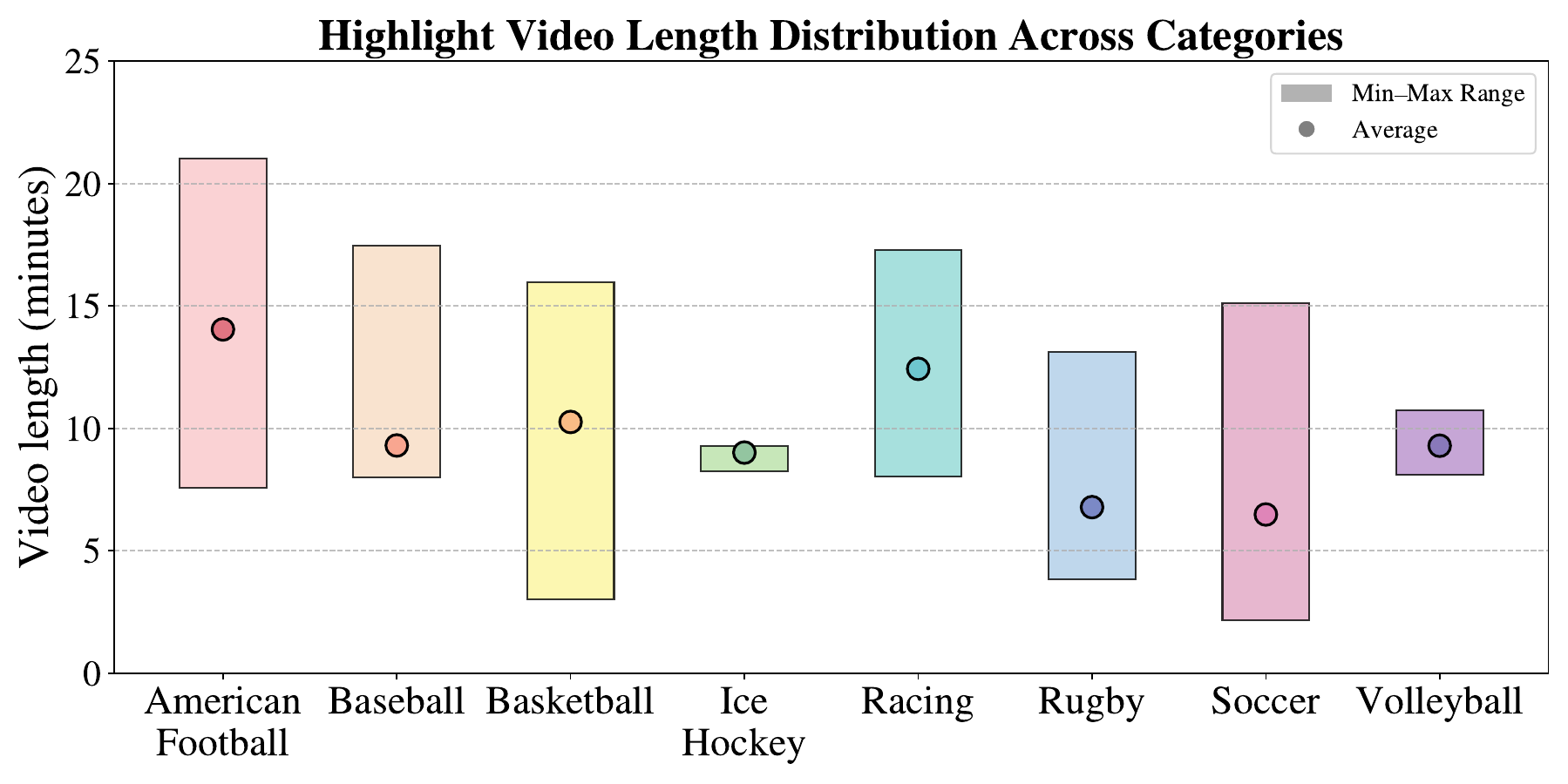}
  \caption{
    Video length distribution across categories. (Top) Full video length. (Bottom) Highlight video length.
  }
  \Description{Two box plots showing the distribution of video lengths across eight sports categories. The top plot shows full video durations ranging roughly from 60 to 200 minutes, while the bottom plot shows highlight video durations ranging roughly from 3 to 30 minutes.}
  \label{fig:statistic}
\end{figure}
\section{SVHighlights}
\label{sec:benchmark}

We introduce \textbf{SVHighlights}, a benchmark specifically designed for highlight detection in extremely long sports videos. This section describes the dataset construction process, including data collection, video trimming, highlight alignment, and label generation.

\subsection{Dataset Collection}

We collected a total of 320 full-length videos from official YouTube channels, with an average duration of 2.00 hours and a cumulative duration of 640.18 hours. As shown in Table~\ref{tab:highlight-benchmarks}, our dataset comprises significantly longer videos compared to existing video highlight detection datasets. This makes it particularly well-suited for evaluating methods designed for long-form highlight detection. As shown in Figure~\ref{fig:statistic}, the videos in our dataset span eight different sports—American football, baseball, basketball, ice hockey, racing, rugby, soccer, and volleyball. Each sport is designed to have an equal number of videos, with 40 videos per category. This diversity in both video length and subject matter enhances the richness and versatility of our dataset compared to existing benchmarks. Moreover, the analysis of full-length and highlight duration distributions reveals that highlight length is not simply proportional to the full video duration but rather depends on the underlying content, indicating that our dataset effectively reflects the nature of real-world highlights. To ensure reliable highlights, we selected only video pairs where the highlight video is sourced from the same broadcast as the full-length video, ensuring consistent visual and audio content. Additionally, we included only those with more than 10,000 views that were produced by neutral sports associations or leagues, rather than by specific teams.

\subsection{Video Trimming}
Full-length videos often include segments unrelated to the target game, such as introductions, footage from previous matches, and post-game interviews. To ensure accurate evaluation, we manually trimmed the full-length videos to retain only the actual game footage, removing unrelated segments at the beginning and end. The same rule applies to every sport---only the cues marking the game start and end differ, as listed in Appendix Table~\ref{tab:trimming}---and requires only one boundary judgment per video, not per-clip annotation. All content within the target game---including timeouts, halftime breaks, and replays---is fully preserved. We did not trim the highlight videos, as subsequent filtering steps remove any highlight segments not present in the full video.

\begin{figure}[t]
  \centering
  \includegraphics[width=\columnwidth]{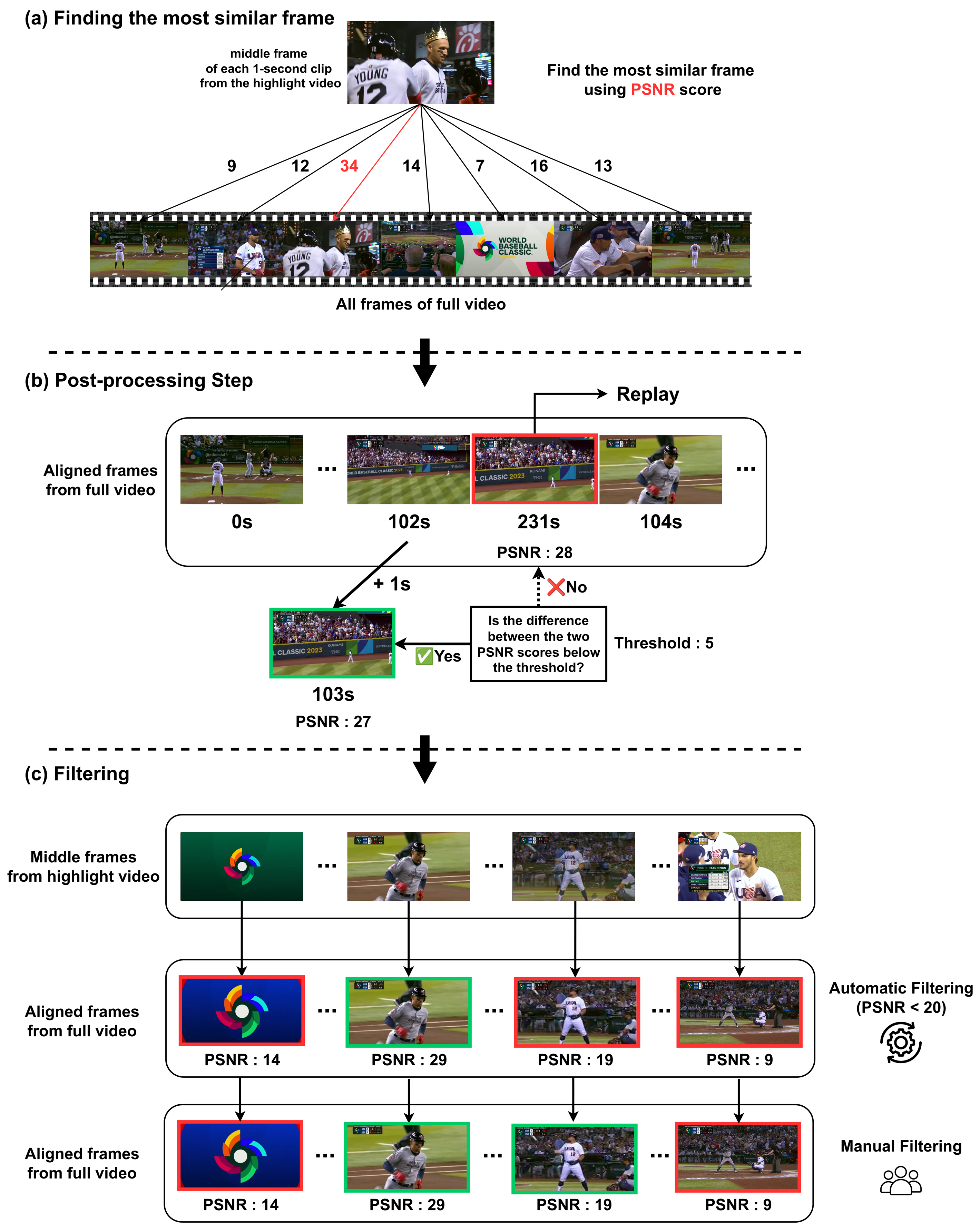}
  \caption{
Overview of the highlight alignment pipeline.
(a) Each highlight frame is aligned to the most similar full-video frame using PSNR.
(b) Temporal consistency is enforced by comparing PSNR differences with a threshold $\tau$.
(c) Automatic PSNR-based filtering followed by manual refinement removes mismatched frames.
}
  \Description{Diagram of the three-stage highlight alignment pipeline: PSNR-based frame matching, temporal post-processing with a threshold, and automatic plus manual filtering.}
  \label{fig:pipeline}
\end{figure}

\subsection{Highlight Alignment Algorithm}

In previous works, benchmarks were built by hiring annotators to assign saliency scores to video clips or shots. However, this method has two major drawbacks: (1) It is both time-consuming and expensive to score every part of long videos. (2) The annotators may not have sufficient domain expertise for the diverse range of content. By contrast, our approach uses official highlight videos produced by professional broadcasters as ground truth, which naturally addresses both issues. We propose a highlight alignment algorithm that identifies and marks segments in full-length videos that appear in the highlight videos. An overview of this highlight alignment algorithm is illustrated in Figure~\ref{fig:pipeline}.

\subsubsection{Finding the Most Similar Frame} Comparing every frame between the full video and the highlight video is extremely time-consuming. Therefore, we first downsample the resolution of the videos to 144p. For alignment, we use all frames from the full video, but only the middle frame of each 1-second clip from the highlight video.
To compare frames at the pixel level, we use the PSNR score, which is inversely related to the mean squared error—so the higher the PSNR score, the more similar the two images are. For each middle frame in the highlight video, we compute the PSNR scores against all frames in the full video and select the frame with the highest score.

We chose PSNR over learned feature-based approaches (e.g., CLIP~\cite{radford2021learning}, ResNet~\cite{he2016deep}) because long-form sports videos contain many visually similar scenes---repeated plays, recurring camera angles, similar field views---which cause feature-based methods to produce false matches, whereas pixel-level PSNR reliably distinguishes near-identical frames.

\begin{figure*}[t]
  \centering
  \includegraphics[width=\textwidth]{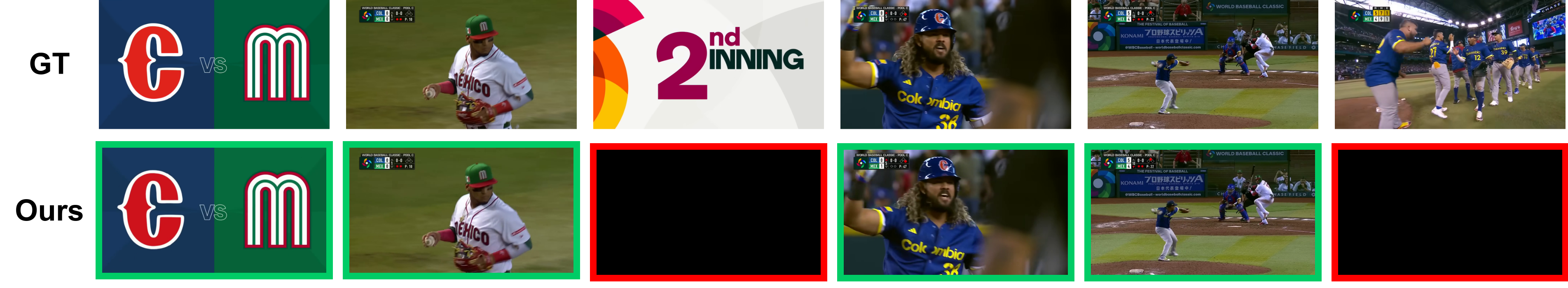}
  \caption{
    \textbf{Example of highlight alignment and filtering results on a baseball video.}
    Each column shows a ground-truth highlight frame (GT, top) and its aligned
    full-video frame (Ours, bottom). Green boxes indicate successful alignments;
    red boxes indicate frames filtered out during the filtering step, with a black frame shown in place of the aligned frame for visualization.
  }
  \Description{Grid of frame pairs comparing ground-truth highlight frames with aligned full-video frames for baseball. Green boxes indicate successful alignments and red boxes indicate filtered frames replaced by black frames.}
  \label{fig:alignment_example}
\end{figure*}

\subsubsection{Post-processing Step} While the above matching works well in most cases, we observed two types of errors: (1) the same full-video frame is repeatedly selected for consecutive highlight frames when they are visually similar, and (2) a replay segment in the full video is matched instead of the actual gameplay moment, since replays are visually identical to the original scenes. Both issues arise because purely PSNR-based matching ignores the temporal order of the video.

To address this, we introduce a post-processing rule based on the following intuition: each highlight clip is a continuous excerpt from the broadcast, so consecutive highlight frames within a clip correspond to temporally adjacent frames in the full video. We therefore prefer the \textit{temporally expected} frame---located one second after the previous match---unless a significantly better match exists elsewhere.
 
Formally, suppose we have already aligned the first $i{-}1$ highlight frames, and let $p$ denote the position of the most recently aligned frame in the full video. For the $i$-th highlight frame $h_i$, we compare two candidates:
\begin{itemize}
\item \textbf{Best match} $f^*$: the frame with the highest PSNR across the entire full video, i.e., $f^* = \arg\max_t \operatorname{PSNR}(h_i, f_t)$
\item \textbf{Expected next} $f^+$: the frame one second after the previous alignment, i.e., $f^+ = f_{p + r}$, where $r$ is the frame rate
\end{itemize}

The aligned frame is then selected as:
\begin{equation}
\operatorname{align}(h_i) =
\begin{cases}
    f^+, & \text{if } \operatorname{PSNR}(h_i, f^*) - \operatorname{PSNR}(h_i, f^+) \leq \tau \\
    f^*, & \text{otherwise}
\end{cases}
\end{equation}

In other words, we default to the temporally expected frame $f^+$ and only switch to the global best match $f^*$ when its PSNR score exceeds that of $f^+$ by more than a threshold $\tau$. This encourages temporally smooth alignments while still allowing jumps when a clearly better match is found at a different position in the full video.

\subsubsection{Filtering} Highlight videos occasionally contain frames that do not appear in the corresponding full videos---for example, sponsor logos, intro/outro sequences, or graphical overlays inserted during scene transitions. As illustrated in Figure~\ref{fig:alignment_example}, such mismatched frames are indicated by red boxes, where no valid alignment exists and the frame is replaced by a black frame. Since these frames can introduce incorrect alignments and undermine the reliability of the benchmark, we introduce a filtering step to remove them.

\paragraph{Automatic filtering.}
Manually inspecting every aligned frame pair would be prohibitively time-consuming, so we first apply an automatic filtering stage. A PSNR score below 20 is commonly regarded as indicating significant perceptual dissimilarity between two images; we therefore remove all aligned pairs whose PSNR falls below this threshold. This step successfully eliminates the majority of mismatched frames.

\paragraph{Manual filtering.}
Although automatic filtering handles most cases, it can produce two types of errors.
\textit{False negatives}: some correctly aligned frames are erroneously removed because editing effects (e.g., fade transitions or overlaid scoreboards) lower their PSNR scores despite depicting the same scene.
\textit{False positives}: in rare cases, frames with high PSNR scores are retained even though they do not correspond to the same scene---for instance, when visually similar but temporally unrelated frames happen to yield high scores.
To correct these errors, two annotators independently inspect aligned pairs in 16-pair grid images (Appendix Fig.~\ref{fig:manual_filtering})---a binary visual check rather than saliency scoring---restoring falsely removed
frames and discarding falsely retained ones. As shown in Table~\ref{tab:manual_filtering}, manual filtering changes only 3.7\% of automatically aligned frames, 95\% of which merely recover false negatives; genuine errors account for just 0.18\% of all frames (99.82\% precision for the automatic stage).

\begin{table}[t]
  \centering
  \caption{Breakdown of changes made by manual filtering to the automatically aligned frames.}
  \begin{tabularx}{\linewidth}{>{\centering\arraybackslash}X c}
    \toprule
    \textbf{Change type after manual filtering} & \textbf{Ratio} \\
    \midrule
    Correct frame index (no change)           & 88.8\% \\
    Both null (no change)                     & 7.5\% \\
    null $\rightarrow$ frame index (recovery) & 3.5\% \\
    frame index $\rightarrow$ null (error)    & 0.1\% \\
    frame index changed (error)               & 0.1\% \\
    \bottomrule
  \end{tabularx}
  \label{tab:manual_filtering}
\end{table}

\begin{table}[t]
  \centering
  \caption{Per-sport alignment quality after filtering, measured by PSNR, SSIM, and CLIP similarity. The remaining rate is the proportion of frames retained after filtering.}
  \resizebox{\linewidth}{!}{%
  \begin{tabular}{lcccc}
    \toprule
    \textbf{Category} & \textbf{PSNR} & \textbf{SSIM} & \textbf{CLIP Sim.} & \textbf{Remain. Rate (\%)} \\
    \midrule
    Amer. Football & 26.64 & 0.872 & 0.969 & 93.7 \\
    Baseball       & 26.28 & 0.835 & 0.943 & 85.8 \\
    Basketball     & 26.31 & 0.868 & 0.957 & 94.7 \\
    Ice Hockey     & 26.63 & 0.856 & 0.973 & 94.6 \\
    Racing         & 27.99 & 0.873 & 0.939 & 92.4 \\
    Rugby          & 27.43 & 0.865 & 0.941 & 87.2 \\
    Soccer         & 26.43 & 0.877 & 0.953 & 90.0 \\
    Volleyball     & 26.23 & 0.874 & 0.961 & 95.3 \\
    \midrule
    All            & 26.74 & 0.865 & 0.955 & 91.7 \\
    \bottomrule
  \end{tabular}
  }
  \label{tab:matching_quality}
\end{table}

\subsection{Labeling Stage}
After highlight alignment, we generate the final highlight labels for the full video. Many existing VTG models~\cite{lei2021detecting, liu2022umt, moon2023query, xu2024mh, lin2023univtg, sun2024tr, liu2024r} assign saliency scores to every 2-second clip. Following prior approaches, we generate labels at 2-second intervals. Specifically, we first obtain the aligned frame indices from the highlight alignment stage and compute the corresponding timestamps in the full video. For each timestamp, we designate a 1-second interval centered on the frame (i.e., 0.5 seconds before and after) as the highlight segment. With this approach, even in the worst case—such as when the middle frame lies on a shot boundary—the maximum alignment error is limited to 0.5 seconds. Next, we divide the full video into non-overlapping 2-second clips and assign a label of 1 to each clip if at least 50\% of its duration overlaps with any highlighted segment; otherwise, the label is set to 0.

\subsection[Dataset Quality Validation]{Dataset Quality Validation}

\subsubsection{Alignment Quality Evaluation}
We report the alignment quality after filtering in Table~\ref{tab:matching_quality}. We measure the similarity between each highlight frame and its aligned full-video frame using three complementary metrics: PSNR, SSIM~\cite{wang2004image}, and CLIP similarity~\cite{radford2021learning}. The average scores of 26.74, 0.865, and 0.955, respectively, confirm that the retained alignments are highly accurate. Moreover, only 8.3\% of frames are removed during filtering, indicating that the preceding alignment algorithm already produces reliable matches and the filtering step serves primarily as a safeguard against a small number of edge cases.

\subsubsection{Agreement between $f^*$ and $f^+$} Although the post-processing step prefers the temporally expected frame $f^+$, this choice rarely departs from the global best match $f^*$: the frame-index gap between $f^+$ and $f^*$ is below 30 frames ($\approx$1\,s) for 90.0\% of frames, and for at least 79.3\% in every sport (Table~\ref{tab:temporal_order}). That is, the temporally expected frame is almost always nearly identical to the global best match. When $f^*$ and $f^+$ do disagree, $f^*$ is still chosen 51.7\% of the time, so the algorithm does not blindly follow temporal order.

\begin{table}[t]
  \centering
  \caption{Per-sport distribution of frame-index gaps between the global best match $f^*$ and the temporally expected frame $f^+$. Most matches fall within $\approx$1\,s ($<$30 frames), confirming that $f^+$ closely agrees with $f^*$.}
  \resizebox{\linewidth}{!}{%
  \begin{tabular}{lccc}
    \toprule
    \textbf{Sport} & \textbf{$<$30 frames} & \textbf{30--999 frames} & \textbf{$\geq$1000 frames} \\
    \midrule
    Amer. Football & 90.5\% & 2.3\% & 7.2\% \\
    Baseball       & 79.3\% & 9.7\% & 11.0\% \\
    Basketball     & 89.4\% & 3.9\% & 6.6\% \\
    Ice Hockey     & 91.9\% & 2.5\% & 5.5\% \\
    Racing         & 93.7\% & 0.8\% & 5.5\% \\
    Rugby          & 88.5\% & 4.3\% & 7.3\% \\
    Soccer         & 93.2\% & 2.8\% & 3.8\% \\
    Volleyball     & 92.0\% & 3.0\% & 5.1\% \\
    \midrule
    \textbf{Total} & \textbf{90.0\%} & \textbf{3.4\%} & \textbf{6.5\%} \\
    \bottomrule
  \end{tabular}}
  \label{tab:temporal_order}
\end{table}

\subsubsection{Threshold $\tau$} We ablate $\tau$ (Table~\ref{tab:tau_ablation}). A small $\tau=1$ almost always picks $f^*$ (72.3\%), ignoring temporal cues, while a large $\tau=10$ over-relies on $f^+$ (72.6\%) and drops the remaining-frame rate to 80.36\%. We adopt $\tau=5$ as a balanced choice (51.7\% vs.\ 48.3\% selection, 88.16\% remaining rate).

\begin{table}[t]
  \centering
  \caption{Ablation of the post-processing threshold $\tau$. Remaining Rate is the proportion of frames retained after filtering; $f^*$/$f^+$ Ratio denotes the fraction of frames for which each candidate is selected.}
  \begin{tabular}{cccc}
    \toprule
    \textbf{$\tau$} & \textbf{Remaining Rate (\%)} & \textbf{$f^*$ Ratio} & \textbf{$f^+$ Ratio} \\
    \midrule
    1  & 89.78 & 72.3\% & 27.7\% \\
    5  & 88.16 & 51.7\% & 48.3\% \\
    10 & 80.36 & 27.4\% & 72.6\% \\
    \bottomrule
  \end{tabular}
  \label{tab:tau_ablation}
\end{table}

\subsubsection{Label Quality User Study} To validate the quality of our automatically generated labels, we conducted a user study. Ten participants rated 120 clips (15 per sport) on a 1--5 Likert scale, drawn equally from three groups: positive clips from ground-truth highlights, near-boundary negatives within 30\,s of a highlight boundary, and far negatives beyond 30\,s. Positive clips averaged 3.42 versus 1.94 and 1.97 for near- and far-boundary negatives, showing that the labels agree well with human judgment; the nearly identical near- and far-negative scores further indicate minimal label noise around highlight boundaries. Agreement also varies by sport (Table~\ref{tab:user_study}): those with clear cues (e.g., Soccer, Basketball) show large positive--negative gaps and high inter-rater agreement (Krippendorff's $\alpha=0.65$--$0.72$), whereas sports requiring domain knowledge (e.g., Racing, Rugby) show small gaps and low agreement ($\alpha=0.22$ for Racing), supporting our use of professionally edited highlights as ground truth.

\begin{table}[t]
  \centering
  \caption{Per-sport user study results for label quality validation. Diff is the positive$-$negative mean-score gap; $\alpha$ is Krippendorff's inter-rater agreement.}
  \resizebox{\linewidth}{!}{%
  \begin{tabular}{lcccc}
    \toprule
    \textbf{Sport} & \textbf{Pos. Mean} & \textbf{Neg. Mean} & \textbf{Diff} & \textbf{Krippendorff's $\alpha$} \\
    \midrule
    Soccer         & 4.00 & 1.40 & 2.60 & 0.702 \\
    Basketball     & 3.98 & 1.73 & 2.25 & 0.654 \\
    Amer. Football & 3.84 & 2.14 & 1.70 & 0.530 \\
    Ice Hockey     & 3.36 & 1.92 & 1.44 & 0.459 \\
    Baseball       & 3.08 & 1.66 & 1.42 & 0.644 \\
    Volleyball     & 3.38 & 2.13 & 1.25 & 0.717 \\
    Racing         & 2.58 & 2.05 & 0.53 & 0.221 \\
    Rugby          & 3.10 & 2.58 & 0.52 & 0.527 \\
    \bottomrule
  \end{tabular}}
  \label{tab:user_study}
\end{table}
\begin{figure*}[t!]
  \centering
  \includegraphics[width=\textwidth]{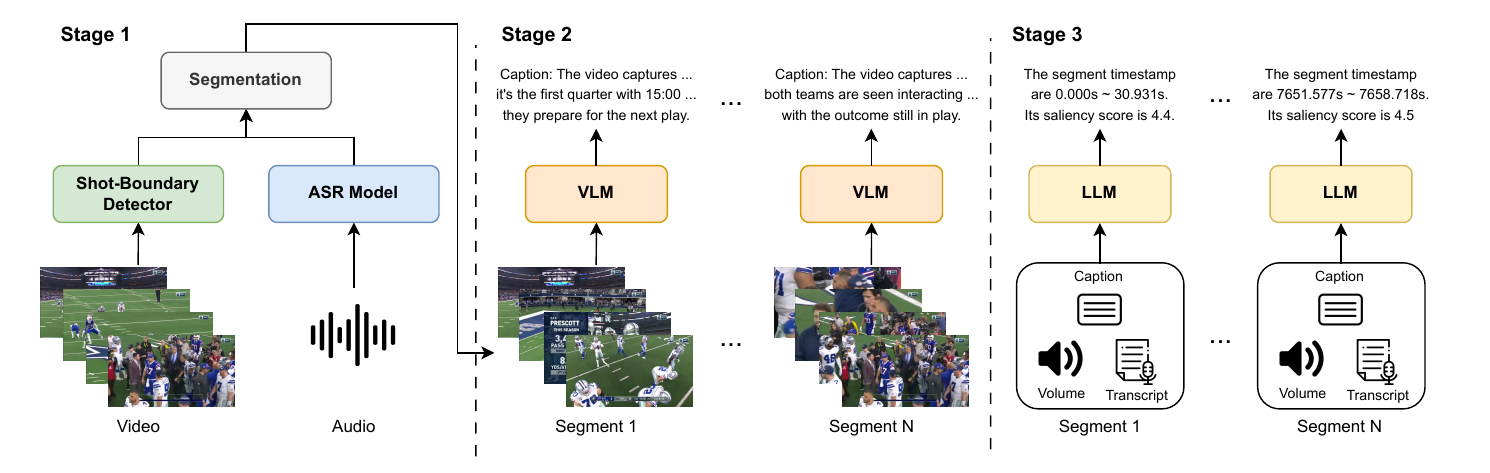}
  \caption{Overview of the TF-SELECTOR framework. Stage~1 (Context-aware segmentation): Shots are detected by a shot-boundary detector, and adjacent shots that share the same content are merged into context-aware segments using transcript information from an ASR model. Stage~2 (Segment captioning): A vision-language model (VLM) generates a caption for each segment. Stage~3 (Segment-level scoring): The LLM predicts a saliency score for each segment using the segment caption, audio volume, and transcript. The predicted score is then assigned to all clips within the corresponding segment.}
  \Description{Block diagram of the TF-SELECTOR framework with three stages: Stage~1 performs context-aware segmentation by detecting shot boundaries and merging adjacent shots using ASR transcripts; Stage~2 generates segment captions using a VLM; Stage~3 uses an LLM to predict saliency scores from captions, audio volume, and transcripts, then assigns scores to individual clips.}
  \label{fig:method}
\end{figure*}

\section{TF-SELECTOR}
\label{sec:method}
In this section, we introduce our training-free framework for long-form video highlight detection, which consists of three stages, as shown in Figure~\ref{fig:method}. In Stage~1, the video is divided into contextually coherent segments. In Stage~2, a VLM generates textual descriptions for each segment, and in Stage~3, an LLM predicts segment-level saliency scores, which are assigned to individual clips.

Since TF-SELECTOR requires no task-specific training, it can be readily applied to long-form videos in a zero-shot manner. Its modular design also allows different VLMs and LLMs to be flexibly substituted, enabling the framework to directly benefit from advances in foundation models.

\subsection{Context-aware Video Segmentation}
\label{subsec:context_segment}
Since it is impractical to process an entire long video at once due to computational cost and context length limitations, the goal of Stage~1 is to divide videos into contextually coherent segments to enable efficient processing using an LLM. To achieve this, we divide the video into shots and then merge adjacent shots based on semantic consistency to generate segments, which serve as the basic units for highlight detection in our approach.

To identify individual shots, we apply a shot-boundary detector that groups visually similar consecutive frames. However, this method relies solely on visual similarity and may split semantically continuous content. For example, a camera angle change may trigger a new shot even when the same scene is still being depicted. As a result, individual shots are often insufficient as semantic units.

To merge shots into semantically complete segments, we use transcript information from an ASR model, which provides word-level timestamps. If the time gap between consecutive words is less than one second, we consider them part of the same sentence. When such a sentence spans two adjacent shots, we treat it as evidence that both shots share the same content and merge them into one segment. To prevent segments from becoming too long, we impose a maximum segment length constraint: if merging two shots would exceed this limit, we do not merge them.

\subsection{Segment Captioning}
\label{captioning}
Since LLMs are unable to understand visual information, we introduce Stage~2 to convert visual content into textual descriptions using a VLM. We first divide the entire video into 2-second clips and sample one frame from each clip. Since each segment from Stage~1 spans multiple such clips, we group the sampled frames by their corresponding segment based on timestamps. These frames are then provided as visual inputs to the VLM, which is prompted with \textit{"Please describe this segment"} to generate a caption for each segment. The resulting caption is used as input to the LLM in Stage~3.

\subsection{Segment-Level Score Prediction}
In Stage~3, we leverage an LLM to predict a saliency score for each segment. Since all clips within a segment share the same context (as ensured by Stage~1), segment-level prediction naturally yields consistent scores across clips depicting the same scene.

The LLM utilizes segment-level information from three modalities: (1) the transcript obtained using an ASR model in Stage~1, (2) the segment caption generated in Stage~2, and (3) the audio volume extracted from the corresponding segment interval. These modalities provide complementary information for score prediction. The caption reflects visual content, the volume captures auditory cues such as crowd reactions or emphasis in commentary, and the transcript conveys linguistic information describing the game and ongoing events.

Once the saliency score is predicted for each segment, it is used to assign scores to each clip. Since each clip has a fixed duration of 2 seconds, while segment boundaries are not aligned with clip boundaries, a clip may overlap with multiple segments. In such cases, we compute the clip-level score using a weighted sum of the scores of overlapping segments, where the weight for each segment is defined as the ratio of its temporal overlap with the clip. The clip-level score $s_C$ is given by:

\begin{equation}
s_C=\sum_{i} \left( \frac{\operatorname{overlap}(C, S_i)}{L_C} \times s_{S_i} \right)
\end{equation}

where $\operatorname{overlap}(C, S_i)$ denotes the temporal overlap between clip $C$ and the $i$-th overlapping segment $S_i$, and $L_C$ is the length of clip $C$. This weighted-sum-based approach allows each clip to be assigned a score that appropriately integrates the saliency of all overlapping segments.

\begin{table*}[t]
\centering
\caption{Zero-shot performance on SVHighlights. V: Video, A: Audio. The best results are highlighted in bold, and the second best are underlined.}
\begin{tabular}{llccccc}
\toprule
\textbf{Method} & \textbf{Input} & \textbf{Scoring} & \textbf{mAP} & \textbf{HIT@1} & \textbf{HIT@K} & \textbf{IoU} \\ \midrule
\rowcolor{gray!15} \rule{0pt}{10pt} \textit{VTG-tuned Non-LLMs} &  &  &  &  &  &  \\ 
Moment-DETR & V & Clip & 8.94 & 5.00 & 7.10 & 3.76 \\
UMT & V+A & Clip & 11.30 & 13.75 & 11.77 & 6.40 \\
QD-DETR & V & Clip & 11.09 & 7.81 & 10.92 & 6.17 \\
MH-DETR & V & Clip & 8.75 & 2.50 & 4.94 & 2.68 \\
UniVTG & V & Clip & 9.29 & 3.12 & 6.33 & 3.34 \\
TR-DETR & V & Clip & 12.72 & \underline{24.06} & \underline{12.86} & 7.09 \\
CG-DETR & V & Clip & 10.36 & 10.31 & 9.74 & 5.37 \\ \midrule
\rowcolor{gray!15} \rule{0pt}{10pt} \textit{VTG-tuned Non-LLMs with pre-training} &  &  &  &  &  &  \\ 
Moment-DETR & V & Clip & 9.23 & 4.69 & 7.31 & 3.90 \\
UMT & V+A & Clip & 10.11 & 7.81 & 8.64 & 4.66 \\
UniVTG & V & Clip & 10.26 & 2.81 & 8.96 & 4.78 \\
CG-DETR & V & Clip & 11.30 & 13.44 & 6.73 & 3.91 \\ \midrule
\rowcolor{gray!15} \rule{0pt}{10pt} \textit{Segment-based Non-LLM} &  &  &  &  &  &  \\
SL-Module & V & Segment & 8.82 & 5.62 & 5.99 & 3.14 \\ \midrule
\rowcolor{gray!15} \rule{0pt}{10pt} \textit{VTG-tuned Vid-LLMs} &  &  &  &  &  &  \\
VTG-LLM & V & Clip & 11.64 & 23.44 & 9.40 & \underline{7.63} \\
TimeChat & V & Clip & 12.40 & 14.69 & 9.42 & \underline{7.63} \\
TRACE & V & Clip & \textbf{23.14} & 23.12 & 9.38 & \underline{7.63} \\ \midrule
\rowcolor{gray!15} \rule{0pt}{10pt} \textit{Training-free Approach} &  &  &  &  &  &  \\
TF-SELECTOR (Ours) & V+A & Segment & \underline{12.81} & \textbf{26.56} & \textbf{16.90} & \textbf{10.58} \\ \bottomrule
\end{tabular}
\label{tab:main_results}
\end{table*}

\section{Experiments}
\label{sec:experiments}

\subsection{Experimental Setup}
\subsubsection{Implementation Details} We set the predefined alignment threshold $\tau$ to 5 in our alignment algorithm, and the maximum segment length to 2 minutes. For shot-boundary detection, we use TransNet V2~\cite{soucek2024transnet}, and WhisperX-large-v2~\cite{bain2022whisperx} is employed as the ASR model. For segment captioning, we use InternVL2.5-8B~\cite{chen2024expanding} as the VLM, and Llama-3-8B~\cite{grattafiori2024llama} serves as the LLM for segment-level saliency prediction. All experiments were conducted using a single NVIDIA A6000 GPU.

\subsubsection{Baselines} We compare our method with three types of baselines: (1) \textbf{VTG-Tuned Non-LLMs}, including Moment-DETR~\cite{lei2021detecting}, UMT~\cite{liu2022umt}, QD-DETR~\cite{moon2023query}, MH-DETR~\cite{xu2024mh}, UniVTG~\cite{lin2023univtg}, TR-DETR~\cite{sun2024tr}, and CG-DETR~\cite{moon2023correlation}, which are transformer-based models fine-tuned on QVHighlights~\cite{lei2021detecting}, with some variants additionally pre-trained.
(2) \textbf{Segment-based Non-LLM}, namely SL-Module~\cite{xu2021cross}, which we include as the only segment-based highlight detection method with publicly available code.
(3) \textbf{VTG-Tuned Vid-LLMs}, including VTG-LLM~\cite{guo2025vtg}, TimeChat~\cite{ren2024timechat}, and TRACE~\cite{guo2024trace} with 7B LLMs, all of which are fine-tuned on a video-centric instruction-tuning dataset.
In contrast, our TF-SELECTOR requires no task-specific training on any highlight detection dataset; it operates in a fully zero-shot manner by leveraging off-the-shelf foundation models. Since both VTG-tuned categories are designed for Video Temporal Grounding (VTG), we provide the following query as input: \textit{"Highlight of this \{video\_type\} video"}.

\subsubsection{Evaluation Metrics} Following~\cite{sun2014ranking}, we use mean average precision (mAP) and HIT@1. However, as shown in Figure~\ref{fig:statistic}, the number of highlight clips in SVHighlights varies substantially not only across sports but also across videos within the same sport, reflecting the inherent diversity of real-world highlights. Ranking-based metrics such as mAP and HIT@1 evaluate the quality of ranked predictions, with an emphasis on whether relevant segments are ranked highly, but they do not capture how well a model covers the full extent of highlights whose length differs significantly from video to video. To address this, we additionally introduce HIT@K and IoU, leveraging the clear ground truth provided by official highlight videos. HIT@K measures the proportion of ground-truth clips captured in the top-K predictions, where K equals the number of ground-truth highlight clips for each video, thereby adapting to the variable highlight length. IoU quantifies the temporal overlap between predicted and ground-truth clips, providing a holistic measure of detection quality regardless of highlight duration.

\subsection{Results}
Table~\ref{tab:main_results} presents the results of existing VTG baselines and our TF-SELECTOR on SVHighlights. Experimental results show that TF-SELECTOR outperforms the second-best models by significant margins: +2.50 in HIT@1, +4.04 in HIT@K, and +2.95 in IoU. This demonstrates the effectiveness of our approach, which predicts segment-level saliency scores using segment captions, transcripts, and audio volume as inputs. Although TF-SELECTOR ranks second in mAP, this is mainly because TRACE tends to assign scores to only a small subset of clips while assigning zero to the rest, achieving high accuracy on those few predicted clips. As a result, TRACE attains a high mAP but much lower HIT@K and IoU scores. In addition, SL-Module, the only other segment-based baseline, performs poorly across all metrics: its fixed-length segments, designed for short videos, fail to capture the temporal dynamics of hour-long broadcasts, whereas TF-SELECTOR's context-aware, variable-length segments scale effectively to long-form highlight detection.

\subsection{Ablation Study}
\subsubsection{Effect of VLM} Table~\ref{tab:ablation_captioner} reports an ablation study on different segment captioners. Among the evaluated captioners, LLaVA-OV-7B exhibits the lowest performance across all metrics. Qwen2.5-VL-7B achieves higher scores than InternVL2.5-8B in mAP and HIT@1, indicating stronger precision on the top-ranked clips. By contrast, InternVL2.5-8B demonstrates superior performance in HIT@K and IoU, which evaluate the overall coverage and consistency of highlight localization. As mentioned earlier, mAP and HIT@1 mainly reflect ranking precision rather than overall highlight coverage. Since our goal is to ensure consistent and comprehensive coverage of highlights with varying durations across long videos, rather than focusing solely on ranking precision, we adopt InternVL2.5-8B as the main VLM for TF-SELECTOR. The performance gap across different captioners remains relatively small, indicating that the choice of captioner has only a minor impact on the overall performance.

\subsubsection{Effect of LLM} Table~\ref{tab:ablation_llm} compares four open-source LLMs for segment-level saliency prediction. Llama3-8B consistently outperforms the other models across all metrics, achieving improvements of +3.22 in mAP, +12.81 in HIT@1, +5.56 in HIT@K, and +2.90 in IoU over Llama2-7B. This suggests that more recent LLMs with improved instruction-following capabilities are better at predicting saliency from multimodal segment descriptions. These results demonstrate that the choice of LLM plays a significant role in saliency estimation within our framework.

\begin{table}[t!]
\centering
\caption{Ablation study on the effect of different captioners (VLMs). The best results are highlighted in bold, and the second best are underlined.}
\resizebox{\linewidth}{!}{%
\begin{tabular}{lcccc}
\toprule
\textbf{Captioner} & \textbf{mAP} & \textbf{HIT@1} & \textbf{HIT@K} & \textbf{IoU} \\ \midrule
LLaVA-OV-7B & 12.42 & 25.31 & 15.28 & 9.86 \\
Qwen2.5-VL-7B & \textbf{13.54} & \textbf{28.12} & \underline{16.72} & \underline{10.09} \\
InternVL2.5-8B & \underline{12.81} & \underline{26.56} & \textbf{16.90} & \textbf{10.58} \\ \bottomrule
\end{tabular}
}

\label{tab:ablation_captioner}
\end{table}

\begin{table}[t!]
\centering
\caption{Ablation study on the effect of different LLMs. The best results are highlighted in bold, and the second best are underlined.}
\resizebox{\linewidth}{!}{%
\begin{tabular}{lcccc}
\toprule
\textbf{LLM} & \textbf{mAP} & \textbf{HIT@1} & \textbf{HIT@K} & \textbf{IoU} \\ \midrule
Llama2-7B & 9.59 & 13.75 & 11.34 & 7.68 \\
Qwen2.5-7B & 9.90 & \underline{21.88} & 13.80 & 9.53 \\
Mistral-7B & \underline{11.05} & 21.25 & \underline{15.14} & \underline{10.02} \\
Llama3-8B & \textbf{12.81} & \textbf{26.56} & \textbf{16.90} & \textbf{10.58} \\ \bottomrule
\end{tabular}
}
\label{tab:ablation_llm}
\end{table}

\begin{table}[t!]
\centering
\caption{Ablation study on the effect of different input modalities. C: Caption, A: Audio Volume, T: Transcript, S: Score. The best results are highlighted in bold, and the second best are underlined.}
\resizebox{\linewidth}{!}{%
\begin{tabular}{lcccc}
\toprule
\textbf{Modality} & \textbf{mAP} & \textbf{HIT@1} & \textbf{HIT@K} & \textbf{IoU} \\ \midrule
(C) $\rightarrow$ S & 10.54 & 15.00 & 12.54 & 8.64 \\
(C, A) $\rightarrow$ S & 10.29 & 16.25 & 12.38 & 9.23 \\
(C, T) $\rightarrow$ S & \underline{12.18} & \underline{21.25} & \underline{16.21} & \textbf{11.11} \\
(C, T, A) $\rightarrow$ S & \textbf{12.81} & \textbf{26.56} & \textbf{16.90} & \underline{10.58} \\ \bottomrule
\end{tabular}
}
\label{tab:ablation_modality}
\end{table}

\subsubsection{Effect of Input Modality} Table~\ref{tab:ablation_modality} presents an ablation study on input modalities. We evaluated four variants: using only captions; captions with audio volume; captions with transcripts; and all three modalities combined. The results show that using only captions and captions plus audio volume yield similar performance, suggesting that audio volume alone does not provide much additional information beyond the captions. However, when transcripts are added instead of audio volume, performance improves across all metrics compared to using captions alone—with HIT@1 increasing by +6.25. This underscores the significant contribution of transcripts for highlight prediction. Finally, the best overall performance is achieved when all three modalities are combined, with the highest mAP, HIT@1, and HIT@K, although (C, T) achieves a slightly higher IoU. Audio volume alone provides limited information beyond captions, but when paired with transcripts, it serves as a reinforcing signal that helps the model more confidently identify highlights.
\section{Conclusion}
\label{sec:conclusion}
In this paper, we introduced SVHighlights, the first highlight detection benchmark for extremely long sports videos exceeding one hour, constructed via a dataset generation pipeline that pairs full-length videos with official highlights. We also proposed TF-SELECTOR, a training-free framework that predicts segment-level highlight scores by integrating visual, textual, and audio modalities, achieving scalability and semantic consistency over long videos. Experiments on SVHighlights show that TF-SELECTOR consistently outperforms state-of-the-art baselines across HIT@1, HIT@K, and IoU, while ablation studies confirm the benefit of multimodal inputs, particularly transcripts. We believe SVHighlights and TF-SELECTOR will foster further research on scalable highlight detection and multimodal reasoning in long-form videos.

\section{Limitations and Future Work}
\label{sec:limitations}
The dataset generation pipeline of SVHighlights requires paired full-length and highlight videos, limiting its extension beyond sports. Additionally, preprocessing steps such as frame alignment and captioning are time-consuming, and the training-free reliance on VLMs and LLMs may constrain flexibility. Future work will focus on expanding the dataset to broader long-form domains, optimizing preprocessing for scalability, and integrating more robust reasoning mechanisms.

\begin{acks}
This work was supported by Institute of Information \& Communications Technology Planning \& Evaluation (IITP) grant funded by the Korea government (MSIT) (No. RS-2022-II220608/2022-0-00608, Artificial Intelligence research about multimodal interactions for empathetic conversations with humans, No. IITP-2026-RS-2024-00360227, Leading Generative AI Human Resources Development, No. RS-2025-25442824, AI Star Fellowship Program (Ulsan National Institute of Science and Technology), \& No. RS-2020-II201336, Artificial Intelligence graduate school support (UNIST)).
\end{acks}

\bibliographystyle{ACM-Reference-Format}
\balance
\bibliography{main}

\appendix
\setcounter{section}{0}
\renewcommand{\thesection}{\Alph{section}}

\section{Prompt Details}

\begin{figure*}[t]
  \centering
  \includegraphics{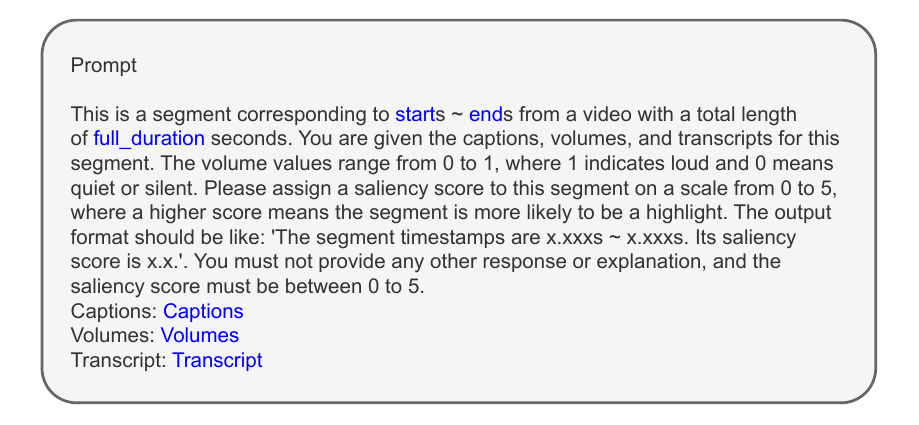}
  \caption{Prompt used for segment-level score prediction.}
  \Description{Text showing the full prompt template provided to the LLM for segment-level saliency score prediction, including system instructions, input format with caption, transcript, and audio volume fields, and output format requesting a score between 0 and 10.}
  \label{fig:prompt_details}
\end{figure*}

We provide the detailed prompt for segment-level score prediction in Figure~\ref{fig:prompt_details}.

\section[Video Trimming Details]{Video Trimming Details}
Without trimming, non-game segments such as pre-game analyses, half-time interviews, commercials, and post-game ceremonies would introduce spurious negatives during alignment, since they have no highlight counterpart yet share the broadcast's visual style. Trimming was performed manually by the authors using the per-sport game start and end cues listed in Table~\ref{tab:trimming}. Each cue was chosen to coincide with an unambiguous, visually or aurally salient event (e.g., the opening whistle, the kickoff, or the first pitch) so that boundaries can be located reliably from a quick scan of the broadcast.

\begin{table}[t]
  \centering
  \caption{Per-sport game start and end cues used for video trimming.}
  \resizebox{\linewidth}{!}{%
  \begin{tabular}{lll}
    \toprule
    \textbf{Sport} & \textbf{Game Start} & \textbf{Game End} \\
    \midrule
    Amer. Football & Opening whistle & Final whistle \\
    Baseball       & Just before first pitch & Just after last out \\
    Basketball     & Opening whistle & Final buzzer \\
    Ice Hockey     & Opening faceoff & Final buzzer \\
    Racing         & Starting signal & Winner's checkered flag \\
    Rugby          & Kickoff whistle & Final whistle \\
    Soccer         & Kickoff whistle & Final whistle \\
    Volleyball     & Opening whistle & Final set match point \\
    \bottomrule
  \end{tabular}}
  \label{tab:trimming}
\end{table}

\section{Manual Filtering Details}
\begin{figure*}[t]
  \centering
  \includegraphics[width=\textwidth]{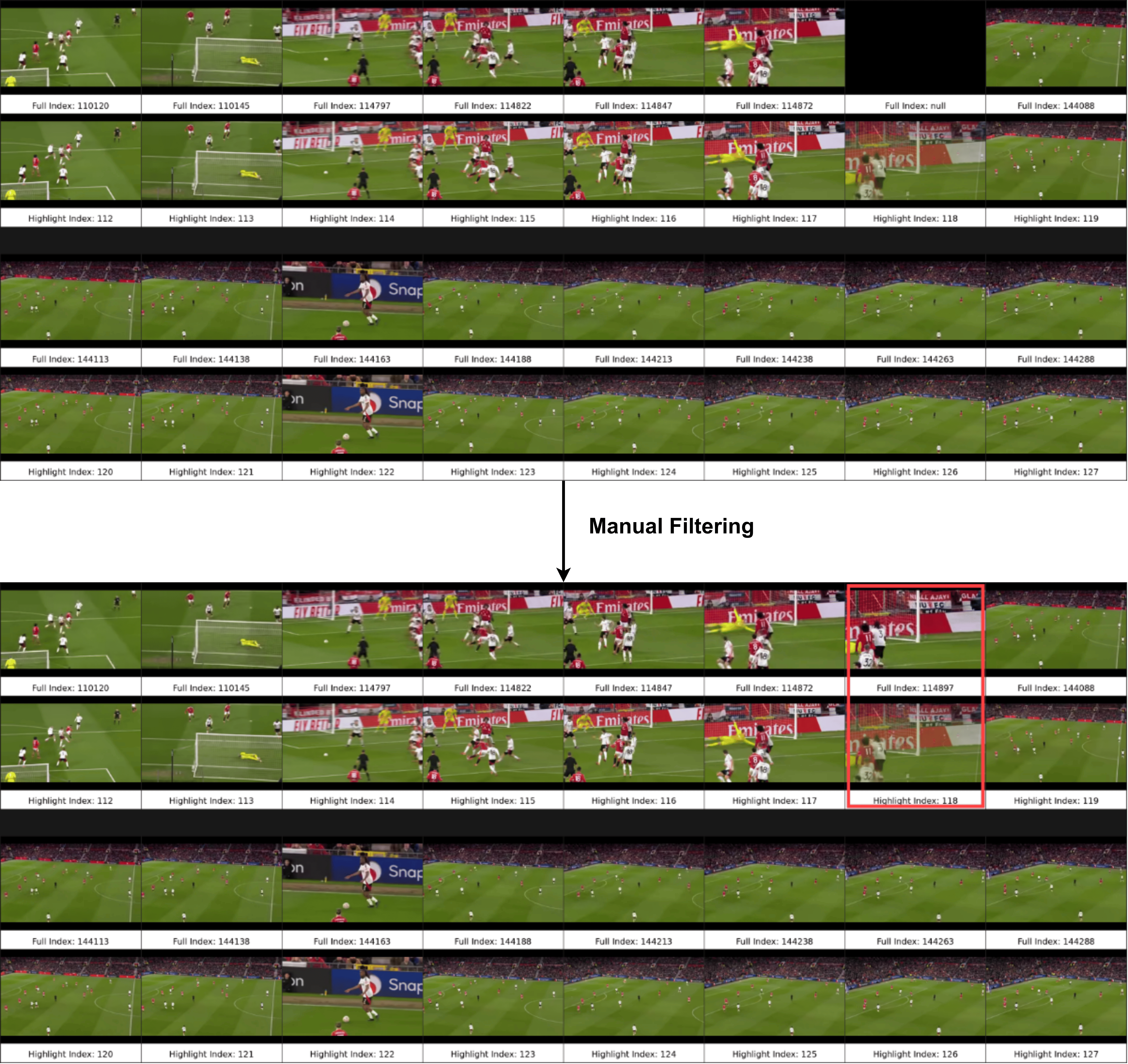}
  \caption{\textbf{An example of the manual filtering process.} We visualized the alignment results as grid images to manually inspect the samples discarded by the automatic filtering. This figure demonstrates a case where a highlight frame was erroneously filtered out due to a low PSNR score caused by frame overlapping during a scene transition. To address such false negatives, we retrieved and verified the original aligned frames. We conducted this manual filtering on the entire SVHighlights dataset to ensure the integrity of the data.}
  \Description{Grid image showing pairs of highlight frames and their aligned full-video frames. One pair is highlighted where a scene transition caused a low PSNR score, leading to erroneous automatic filtering. The original aligned frame is retrieved to correct this false negative.}
  \label{fig:manual_filtering}
\end{figure*}

To correct false negatives from the automatic PSNR filtering stage, we performed an additional manual verification step. For each batch of aligned pairs, we display a 16-pair grid image (Figure~\ref{fig:manual_filtering}) showing the middle frame of each 1-second highlight clip alongside its aligned full-video frame, and the annotator simply judges whether each displayed alignment is correct. This is a binary visual check on existing matches, not a new annotation: annotators do not assign per-clip saliency scores and do not need to watch the original full-length videos. The procedure was carried out by two annotators (the authors), at a cost far lower than the tens to hundreds of hours of per-clip saliency labeling required by existing highlight detection benchmarks. This lightweight verification ensures the precise temporal alignment and overall integrity of the SVHighlights dataset.

\section[Window-based F1 Evaluation]{Window-based F1 Evaluation}
Beyond the ranking- and overlap-based metrics used in the main paper, it is also informative to evaluate highlight detection from the perspective of temporal event detection, where a prediction close to a ground-truth boundary is still practically useful. Inspired by window-based evaluation in the time-series anomaly detection literature~\cite{paparrizos2022volume}, we report an event-level F1: consecutive highlight clips form events, predictions are binarized by top-$k$ selection ($k$ is the number of ground-truth highlight clips), and a predicted event matches a ground-truth event when they overlap within a tolerance window of $w=3$ clips ($\pm6$\,s).
\begin{table}[H]
  \centering
  \caption{Comparison including window-based F1, shown for the strongest baseline on each metric. The best result per column is in bold and the second best is underlined.}
  \resizebox{\linewidth}{!}{%
  \begin{tabular}{lccccc}
    \toprule
    \textbf{Method} & \textbf{mAP} & \textbf{HIT@1} & \textbf{HIT@K} & \textbf{IoU} & \textbf{Window F1} \\
    \midrule
    UMT                & 11.30 & 13.75 & 11.77 & 6.40 & \textbf{30.14} \\
    TR-DETR            & 12.72 & \underline{24.06} & \underline{12.86} & 7.09 & 24.12 \\
    TRACE              & \textbf{23.14} & 23.12 & 9.38 & \underline{7.63} & 17.61 \\
    TF-SELECTOR (Ours) & \underline{12.81} & \textbf{26.56} & \textbf{16.90} & \textbf{10.58} & \underline{27.38} \\
    \bottomrule
  \end{tabular}}
  \label{tab:window_f1}
\end{table}
\noindent Precision, recall, and F1 are then aggregated over all videos. Table~\ref{tab:window_f1} reports this metric alongside the others for the strongest baseline on each metric. TF-SELECTOR attains the second-highest window-based F1 overall (27.38), behind only UMT, and---unlike baselines that excel on only a single metric---remains consistently strong across all metrics, confirming the robustness of its segment-level predictions.

\section{Ethical Considerations}
All videos in SVHighlights were collected from publicly available official YouTube channels operated by professional sports leagues and associations, restricted to videos with over 10,000 views from neutral organizations rather than specific teams. To respect intellectual property, we do not redistribute the original video files and release only video URLs, extracted features, and annotation labels. As the videos are publicly broadcast sports content, they contain no sensitive personal information beyond what is already public, and all annotation was performed by the authors without external crowdworkers. We therefore believe SVHighlights poses minimal ethical risk and is intended solely for advancing research in video understanding.

\end{document}